\documentclass[letterpaper, 10 pt, conference]{ieeeconf}    
\IEEEoverridecommandlockouts    
\usepackage{amsmath}
\usepackage{amssymb}
\usepackage{caption}
\usepackage{cite}

\usepackage{booktabs}
\usepackage{amsmath}
\usepackage{array}
\usepackage{graphicx}
\usepackage{threeparttable}
\usepackage[colorlinks=true,linkcolor=blue,urlcolor=blue]{hyperref}

\allowdisplaybreaks[4]

\title{\LARGE \bf GCNGrasp-VP: Affordance-Guided View Planning for Efficient Task-Oriented Grasping}

\author{Zanjia Tong$^{1}$, Wenlong Dong$^{1}$, Chengjie Zhang$^{1}$, and Hong Zhang$^{1}$ \textit{Life Fellow, IEEE}  
\thanks{$^{1}$Shenzhen Key Laboratory of Robotics and Computer Vision, Southern University of Science and Technology, Shenzhen, China.}}

\usepackage[table]{xcolor}
\definecolor{top1}{rgb}{0.99607843, 0.85098039, 0.38039216}
\definecolor{top2}{rgb}{0.6745098,  0.84313725, 0.55686275}
\definecolor{top3}{rgb}{0.6627451,  0.91372549, 0.89411765}

\begin{document}

\maketitle
\thispagestyle{empty}
\pagestyle{empty}
\bstctlcite{IEEEtran:BSTcontrol}

\begin{abstract}

Task-oriented grasping performance degrades significantly when object views suffer from occlusions. Existing task-oriented grasping methods typically assume task-relevant regions are visible in the initial frame, while view planning approaches enable active perception but often ignore task semantics and rely on time-consuming scene reconstruction. To address these limitations, we present GCNGrasp-VP, an efficient framework integrating affordance field prediction with active view planning. Central to this framework is GCNGrasp-v2, a task-oriented grasp model that simultaneously supports grasp evaluation and affordance field prediction, achieving constant-time inference complexity. Leveraging this capability, our Affordance-guided View Planner (Affordance-VP) utilizes the affordance field as an information gain metric to guide camera observation of task-relevant regions without requiring scene reconstruction. View planning results show that our method significantly outperforms scene-uncertainty-driven baselines with only one view adjustment. Real-world validation further confirms substantial improvements in grasp success rates for single-object scenarios while maintaining millisecond-level computational latency. Code and models are available at \url{https://github.com/Instinct323/GCNGrasp-VP}.

\end{abstract}

\section{INTRODUCTION}

\begin{figure*}[htb]
    \centering
    \includegraphics[width=1.0\linewidth]{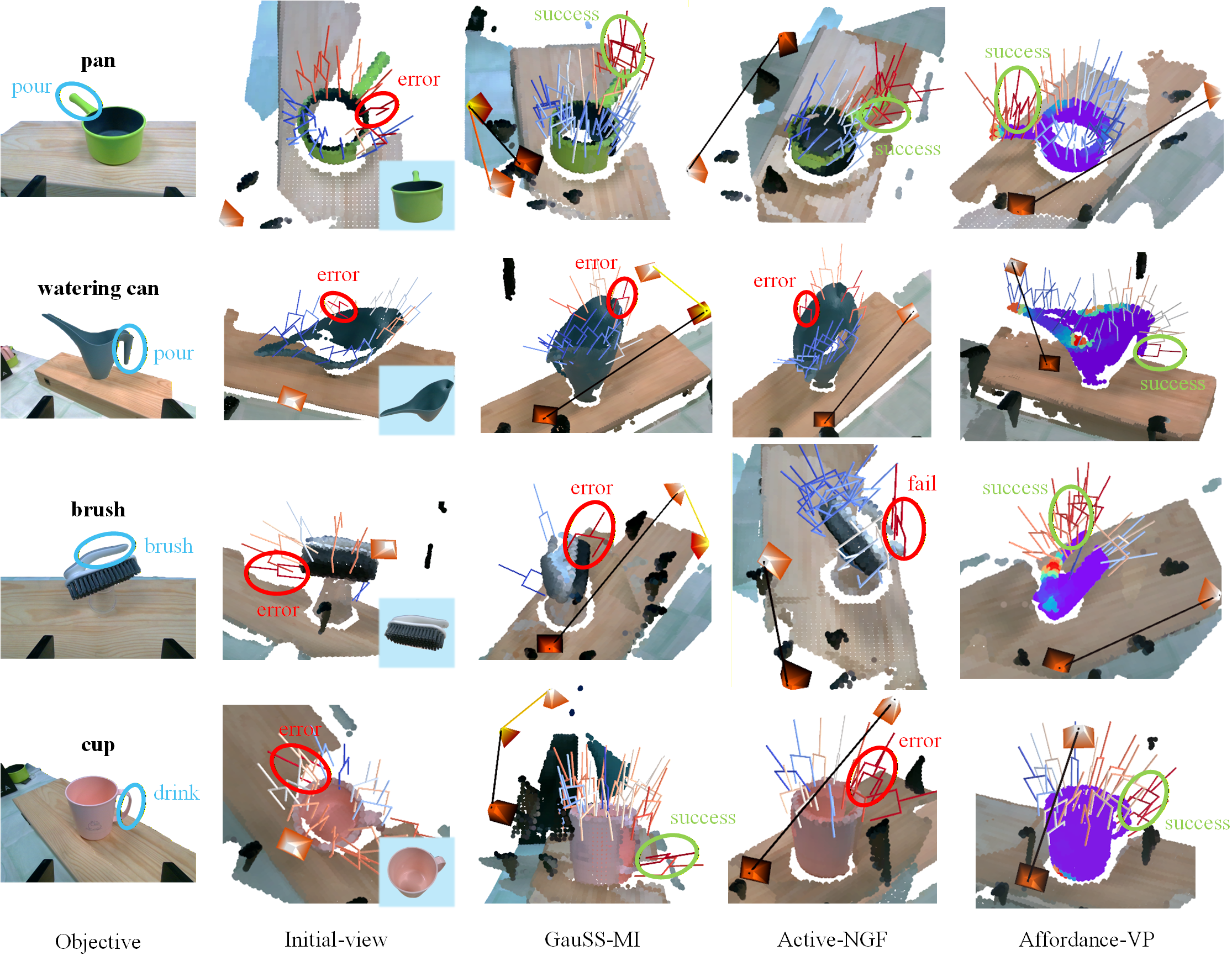}
    \caption{Qualitative comparison of task-oriented grasping results after acquiring additional views using different view planners. Note that GauSS-MI \cite{xie2025GauSSMIGaussianSplatting} requires two additional views due to initialization constraints, whereas other methods require only one. Candidate grasp poses are color-coded by confidence, with warmer colors indicating higher confidence. Circles highlight the grasp pose with the highest confidence, annotated with their final execution outcome (success or error/fail).}
    \label{fig:vis-vp}
\end{figure*}

Task-oriented grasping is a critical component of modular robot manipulation systems, requiring robots to grasp task-relevant regions of objects for manipulation tasks. Unlike task-agnostic grasping, task-oriented grasping must understand the association between geometry and tasks \cite{murali2021SameObjectDifferent}. However, mainstream Task-oriented Grasp (TOG) methods assume the initial view exposes task-relevant regions. In practice, camera positions are arbitrary, and task-relevant regions are often invisible due to self-occlusion or obstacles (Fig.~\ref{fig:vis-vp}). Models \cite{murali2021SameObjectDifferent, tang2023GraspGPTLeveragingSemantic} trained on complete-view datasets like TaskGrasp suffer sharp performance degradation under occlusion. Although large language models \cite{radford2021LearningTransferableVisual, openai2023GPT4TechnicalReport, mirjalili2023LangraspUsingLarge, rashid2023LanguageEmbeddedRadiance, li2024ShapeGraspZeroshotTaskoriented, oort2025OpenvocabularyPartbasedGrasping, liu2025LeveragingSemanticGeometric} or memory retrieval \cite{ju2025RoboABCAffordanceGeneralization, shailesh2025GRIMTaskorientedGrasping, dong2025RTAGraspLearningTaskOriented} enhance semantic understanding, they still rely on passively received initial views. If the initial view lacks task-relevant regions, task-oriented grasping often fails.

Moving the camera enables observing occluded task-relevant regions, yet existing view planning mostly targets task-agnostic grasping, focusing on whole objects instead of specific local regions. Geometry-driven methods \cite{breyer2022ClosedloopNextbestviewPlanning, dai2025ActiveperceptiveLanguageorientedGrasp, liu2026ActiveVLAInjectingActive, shi2025VISOGraspVisionLanguageInformed} only avoid obstacles and cannot guarantee that visible regions contain the required task-relevant regions. Scene-uncertainty-driven methods \cite{gao2024ActivePerceptionGrasp} rely on time-consuming 3D reconstruction, and their view selection based on entropy or reconstruction error is blind to task-oriented grasping, often prioritizing task-irrelevant regions. These limitations leave a gap for solutions that understand task semantics and focus on task-relevant regions in real time.

Our core insight is that TOG model knowledge supports both grasp evaluation and affordance field generation for camera movement. To this end, we propose GCNGrasp-VP, a framework combining affordance prediction and view planning (Fig.~\ref{fig:sys-ov}). GCNGrasp-v2 improves upon GCNGrasp-v1 \cite{murali2021SameObjectDifferent} by using a segmentation-style architecture for simultaneous grasp evaluation and affordance prediction with constant-time inference. Furthermore, Affordance-VP uses the affordance field as an information gain metric to guide the camera toward task-relevant regions without explicit scene reconstruction. The main contributions of this paper are summarized as follows:

\begin{itemize}

\item We propose GCNGrasp-v2, a TOG model that simultaneously supports grasp evaluation and affordance field prediction, achieving constant-time inference complexity while maintaining state-of-the-art performance.

\item We design Affordance-VP, a planner that incorporates the affordance field as a task-aware information gain metric into the view planning loop for the first time, enabling active observation tailored to specific tasks.

\item Experiments demonstrate that our approach significantly outperforms scene-uncertainty-driven baselines with only one view adjustment. Real-world deployments further confirm that our method substantially improves grasp success rates in single-object scenarios with minimal latency.
\end{itemize}

\section{RELATED WORK}

\subsection{Task-Oriented Grasp Model}

Most existing task-oriented grasping methods operate under the strong assumption that the initial observation view sufficiently exposes all task-relevant regions. Benchmark datasets curated under this assumption, such as TaskGrasp \cite{murali2021SameObjectDifferent}, typically provide complete views. Consequently, models trained on such data \cite{murali2021SameObjectDifferent, tang2023GraspGPTLeveragingSemantic} suffer significant performance degradation when facing occlusions or suboptimal initial views. To mitigate this data dependency, some approaches construct grasp knowledge bases that generate grasps by retrieving similar observations. These methods remain limited by initial view quality, as significant appearance variations across views often cause retrieval failures \cite{ju2025RoboABCAffordanceGeneralization, shailesh2025GRIMTaskorientedGrasping, dong2025RTAGraspLearningTaskOriented}. Alternatively, other methods employ open-vocabulary models \cite{radford2021LearningTransferableVisual, openai2023GPT4TechnicalReport, mirjalili2023LangraspUsingLarge, rashid2023LanguageEmbeddedRadiance, li2024ShapeGraspZeroshotTaskoriented, oort2025OpenvocabularyPartbasedGrasping, liu2025LeveragingSemanticGeometric} or affordance models \cite{tang2023TaskorientedGraspPrediction, ma2025GLOVERGeneralizableOpenvocabulary, chen2025EnhancingTaskorientedRobotic} to localize task-relevant regions. Although capable of generating fine-grained task heatmaps, these methods still require task-relevant regions to be structurally visible in the initial frame.

\subsection{View Planning for Grasping}

Existing research on view planning for robotic grasping primarily focuses on improving task-agnostic grasping success rates, with few studies exploring how view selection can directly serve specific task requirements. In cluttered environments, obstacle occlusion critically degrades grasp performance. Many works address this by guiding the camera to unoccluded regions using analytical visibility computation \cite{breyer2022ClosedloopNextbestviewPlanning, dai2025ActiveperceptiveLanguageorientedGrasp, liu2026ActiveVLAInjectingActive} or iterative optimization \cite{shi2025VISOGraspVisionLanguageInformed}. However, these methods are concerned with overcoming occlusion only 
and do not take task constraints into account. Even if an object is not occluded by others, its task-relevant regions may remain invisible due to self-occlusion. Thus, task-oriented grasping requires unoccluded views at the local region level, rather than merely at the instance level.

Another category of methods is driven by scene uncertainty. Active-NGF \cite{gao2024ActivePerceptionGrasp} leverages neural fields \cite{mildenhall2022NeRFRepresentingScenes, johari2023ESLAMEfficientDense} to render novel views and selects them based on graspness uncertainty \cite{wang2021GraspnessDiscoveryClutters}. Other approaches in this category \cite{jiang2025FisherRFActiveView, strong2025NextBestSense, xie2025GauSSMIGaussianSplatting}, though not designed for grasping, also demonstrate effective view selection capabilities. While complete reconstruction helps reveal occluded regions, these methods typically select views based on reconstruction error or information entropy. Such mechanisms are agnostic to task priorities and do not guarantee the visibility of regions critical for executing specific tasks.

\begin{figure*}[htb]
    \centering
    \includegraphics[width=0.9\linewidth]{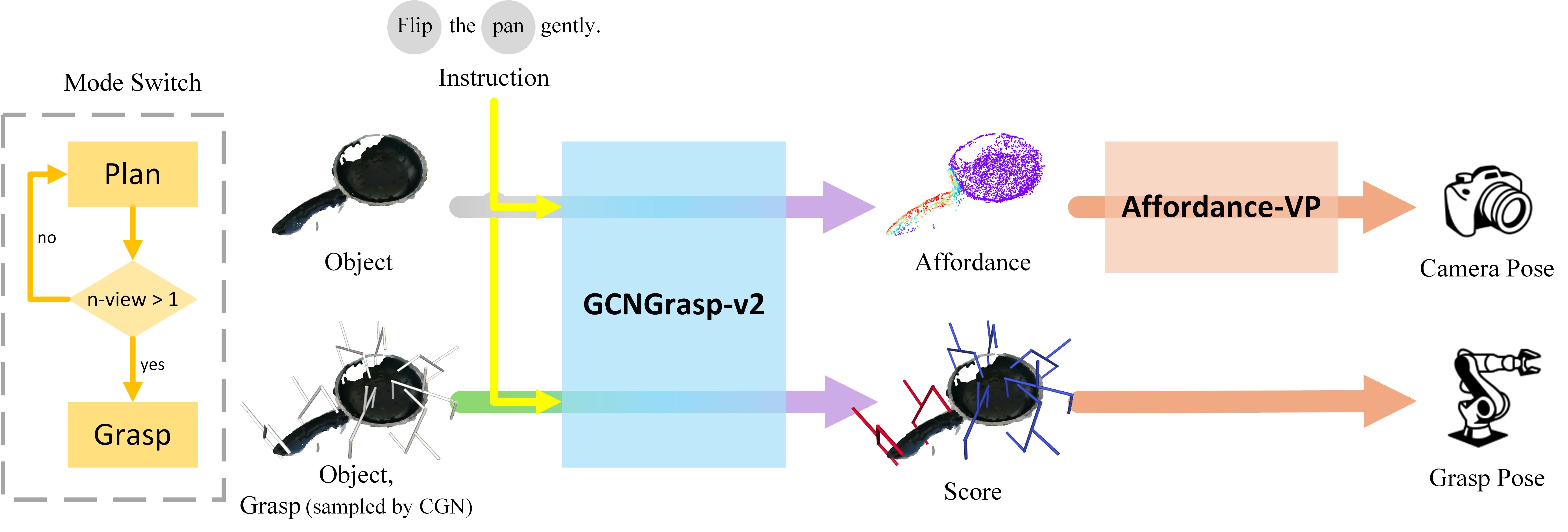}
    \caption{Overview of the GCNGrasp-VP architecture.}
    \label{fig:sys-ov}
\end{figure*}

\subsection{Affordance Field}

Affordance fields are dense scores defined on object point clouds that indicate which regions support specific interactions \cite{deng20213DAffordanceNetBenchmark}. Given their role as robust signals for task-oriented grasping, we explore their application to view planning.

However, existing works utilize the affordance field solely for filtering grasp poses. For instance, GLOVER \cite{ma2025GLOVERGeneralizableOpenvocabulary} identifies high-affordance regions to fit geometric primitives \cite{paschalidou2019SuperquadricsRevisitedLearning} and generate grasps, while others \cite{oort2025OpenvocabularyPartbasedGrasping, song2025Learning6DoFFinegrained, chen2025EnhancingTaskorientedRobotic} use affordance scores to filter out invalid candidate grasps. Methods based on large vision-language models \cite{mirjalili2023LangraspUsingLarge, rashid2023LanguageEmbeddedRadiance, li2024ShapeGraspZeroshotTaskoriented, oort2025OpenvocabularyPartbasedGrasping, liu2025LeveragingSemanticGeometric} follow a similar pattern: they first locate task regions and then search for valid grasps within them. These approaches treat the affordance field solely as a scoring tool for visible regions, limiting its use to the grasp generation phase.

Despite these applications, the potential of the affordance field for guiding view planning remains unexplored. This work is the first to employ the affordance field as an information gain metric within the view planning loop. By guiding the camera toward regions with high affordance scores, our method actively acquires task-oriented observations without requiring time-consuming complete scene reconstruction.

\section{METHOD}

\begin{figure*}[htb]
    \centering
    \includegraphics[width=0.8\linewidth]{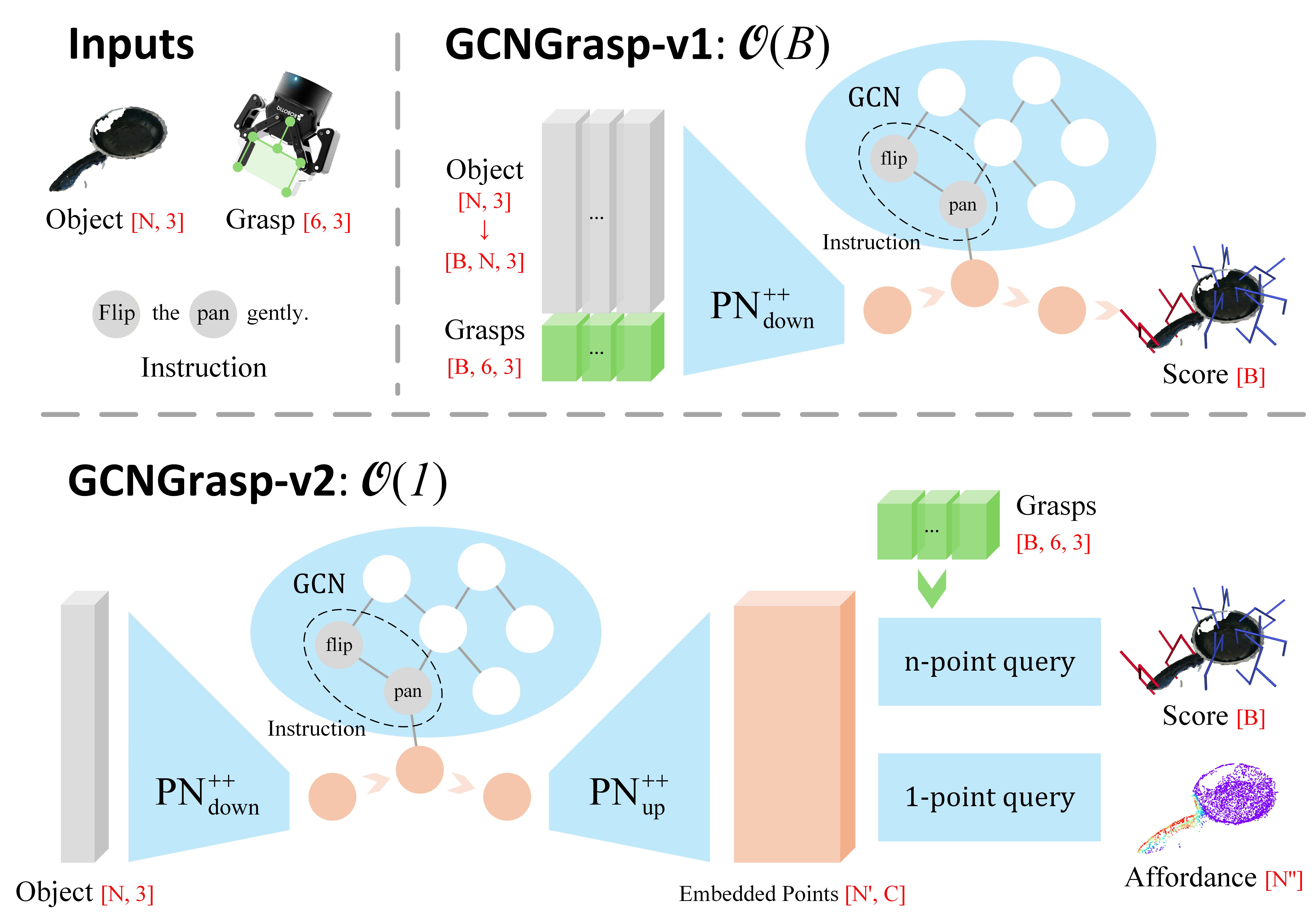}
    \caption{Overview of the GCNGrasp architecture and input definitions, where $B$ denotes the number of candidate grasp poses. \textbf{Top:} GCNGrasp-v1 \cite{murali2021SameObjectDifferent} couples object and grasp features, requiring joint encoding for each candidate. This results in a computational bottleneck with complexity scaling linearly as $\mathcal{O}(B)$. \textbf{Bottom:} GCNGrasp-v2 decouples object-task feature extraction from grasp evaluation. By reusing the global object-task representation, it enables parallel generation of both grasp scores and affordance field, reducing inference complexity to constant time $\mathcal{O}(1)$.}
    \label{fig:arch}
\end{figure*}

\subsection{System Overview}

To enable efficient view planning, we propose the GCNGrasp-VP system, which integrates the TOG model GCNGrasp-v2 with Affordance-VP. Built upon GCNGrasp-v1 \cite{murali2021SameObjectDifferent}, GCNGrasp-v2 retains grasp evaluation capabilities while substantially reducing computational overhead and enabling affordance field prediction. Affordance-VP utilizes this affordance field as an information gain metric to solve the optimal view selection problem (Fig.~\ref{fig:sys-ov}).

\subsection{Task-Oriented Grasp Model}

\newcommand{\pnpp}[1]{\text{PN}^{++}_{\text{#1}}}
\newcommand{\real}[1]{\mathbb{R}^{#1}}
\newcommand{\loss}[1]{\mathcal{L}_{\text{#1}}}

\newcommand{\pcd}{\mathbf{X}}
\newcommand{\pcdctr}{\overline{\mathbf{X}}}
\newcommand{\grasp}{\mathbf{g}}
\newcommand{\grasps}{\mathbf{G}}
\newcommand{\graspsctr}{\overline{\mathbf{G}}}
\newcommand{\affrep}{\mathbf{u}}
\newcommand{\pos}{\mathbf{p}}
\newcommand{\lookat}{\mathbf{v}}

\newcommand{\feat}[1]{\mathbf{h}_{\text{#1}}}
\newcommand{\feats}[1]{\mathbf{F}_{\text{#1}}}
\newcommand{\score}{\hat{y}}
\newcommand{\scores}{\hat{\mathbf{y}}}
\newcommand{\true}{\mathbf{y}}
\newcommand{\aff}{\hat{\mathbf{z}}}
\newcommand{\cluster}{\mathcal{C}}

The effectiveness of view planning depends on the accurate understanding of task-relevant regions by the grasp model. From an existing TOG evaluation model, we introduce affordance supervision signals to equip the model with both grasp scoring and view selection guidance capabilities.

GCNGrasp-v1 employs a classifier-style architecture (Fig.~\ref{fig:arch}). It takes an object point cloud $\pcd \in \real{N \times 3}$, a task instruction $I$, and a single candidate grasp pose $\grasp \in \real{6 \times 3}$ sampled by a task-agnostic grasp model \cite{sundermeyer2021ContactGraspNetEfficient6DoF} as inputs, where $\grasp$ is represented by six control points. The network jointly encodes these inputs into a TOG embedding $\feat{1}$ and produces a binary classification score via a multilayer perceptron (MLP):
\begin{align}
\feat{1} &= \text{GCN}(\pnpp{down}([\pcd, \grasp]), I) \in \real{C} \\
\score &= \text{MLP}(\feat{1}) \in \{0, 1\}
\end{align}
Here, the PointNet++ downsampling network $\pnpp{down}$ \cite{qi2017PointNetDeepHierarchical} extracts geometric features through set abstraction operations, while the graph convolutional network $\text{GCN}$ processes semantic relationships between object categories and tasks in the knowledge graph \cite{jiang2019SemisupervisedLearningGraph, miller1995WordNetLexicalDatabase}.

The affordance field should depend solely on object and task semantics. However, GCNGrasp-v1 tightly couples grasp features with object-task features. This coupling not only prevents affordance field prediction but also incurs a computational bottleneck, as complexity scales linearly with the number of candidate grasps. To address these limitations, we propose GCNGrasp-v2 with a segmentation-style architecture (Fig.~\ref{fig:arch}). This design disentangles object-task features from candidate grasp poses:
\begin{align}
\feat{2} &= \text{GCN}(\pnpp{down}(\pcd), I) \in \real{C} \\
[\pcd', \feats{}] &= \pnpp{up}(\pcd, \feat{2}) \in \real{N' \times (3 + C')}
\end{align}
Here, the PointNet++ upsampling network $\pnpp{up}$ projects the global object-task embedding $\feat{2}$ back onto the high-resolution point cloud $\pcd'$. This process yields the corresponding per-point task-oriented features $\feats{}$.

Leveraging these per-point features, we design a multi-point query mechanism to obtain the TOG embedding $\feat{1}$ for any given $\grasp$. For the six control points of $\grasp$, a KNN-based contact-point query retrieves the neighborhood points $\cluster$. Features within each control point's neighborhood are then aggregated via a group operation, mirroring the Set Abstraction mechanism \cite{qi2017PointNetDeepHierarchical}. Subsequently, an MLP processes the aggregated TOG embedding to produce the compatibility score between the grasp and the task:
\begin{align}
\cluster &= \text{ContactQuery}(\grasp, \pcd', k) \in \mathbb{N}^{6 \times k} \\
\feat{1} &= \text{SA}_{\text{group}}(\cluster, \feats{}) \in \real{6 \times C''} \rightarrow \real{6C''} \\
\score &= \text{MLP}(\feat{1}) \in \{0, 1\}
\end{align}
where $k$ denotes the number of nearest neighbors.

Following established practices \cite{deng20213DAffordanceNetBenchmark, song2025Learning6DoFFinegrained, chen2025EnhancingTaskorientedRobotic}, we generate an affordance field for TOG guidance by decoding the task-oriented features $\feats{}$ using a per-point prediction head. This field is produced via a Set Abstraction operation followed by an MLP, yielding the downsampled point cloud $\pcd''$ and its corresponding affordance scores $\aff''$:
\begin{align}
\label{eq:aff}
[\pcd'', \feats{}'] &= \text{SA}(\pcd', \feats{}) \\
\aff'' &= \text{softmax}(\text{MLP}(\feats{}'))
\end{align}

The improved architecture facilitates the reuse of object-task features, enabling parallel generation of the affordance field and reducing grasp evaluation complexity from $\mathcal{O}(B)$ to $\mathcal{O}(1)$ (Fig.~\ref{fig:arch}).

In the initial view of an object, task-relevant regions are often partially or completely occluded by the object itself (Fig.~\ref{fig:vis-vp}). Consequently, directly localizing target parts as performed by existing methods \cite{deng20213DAffordanceNetBenchmark, ma2025GLOVERGeneralizableOpenvocabulary, song2025Learning6DoFFinegrained, chen2025EnhancingTaskorientedRobotic} is infeasible. We aim to identify regions enriched with TOGs within the visible surface to indirectly target the task-relevant regions. We formalize the supervision label for such a region as a representative point, constructed from the TOG dataset. For each object-task pair, let the set of candidate grasps be $\grasps \in \real{B \times 6 \times 3}$ with center points $\graspsctr \in \real{B \times 3}$ and ground-truth labels $\true \in \{0, 1\}^{B}$. We define the optimal index $j$ as:
\begin{align}
j &= \arg \min_{i} \left\| \graspsctr_i - \frac{\true \cdot \graspsctr}{\sum_i \true_i} \right\| - \left\| \graspsctr_i - \frac{(1 - \true) \cdot \graspsctr}{\sum_i (1 - \true_i)} \right\|
\end{align}
The representative point is then defined as $\affrep = \graspsctr_j$. This strategy selects the point closest to the centroid of positive grasp samples while remaining farthest from the centroid of negative samples, serving as the supervision target for the affordance field.

During the training of GCNGrasp-v2, the model is optimized against TOG labels using a binary cross-entropy loss function, while the representative points constrain the weighted centroid of the affordance field through mean squared error loss:
\begin{align}
\loss{train} &= \loss{cls} + \omega \loss{aff} \\
\loss{cls} &= \frac{1}{B} \sum_{i=1}^B \text{CrossEntropy}(\scores_i, \true_i) \\
\loss{aff} &= \left\| \sum_{i=1}^{N''} \aff''_i \pcd''_i - \affrep \right\|^2
\end{align}
Given that these representative points are approximations derived from statistical distributions, we assign them a small weight $\omega$ as an auxiliary supervision signal. This strategy facilitates the derivation of affordance field prediction capabilities while mitigating the risk of noisy estimates dominating gradient updates.

\subsection{Affordance-Guided View Planner}

Constrained by gravity, the feasible view space is restricted to a compact hemispherical manifold above the object \cite{breyer2022ClosedloopNextbestviewPlanning, shi2025VISOGraspVisionLanguageInformed}. In this domain, only a few views suffice to fully comprehend the object, endowing the next best view problem with excellent convergence. Leveraging this low-entropy property, we employ a greedy strategy to select a target region and optimize its visibility.

After GCNGrasp-v2 outputs the affordance field on the downsampled point cloud (Eq.~\ref{eq:aff}), we upsample the predictions to the original resolution of $\pcd$ to obtain $\aff$. Subsequently, we filter high-confidence points with scores exceeding the 90th percentile of $\aff$, cluster them using DBSCAN \cite{esterDensitybasedAlgorithmDiscovering}, and select the largest cluster $\cluster^*$ as the target region:
\begin{align}
\aff &= \text{upsamp}(\pcd, \pcd'', \aff'') \\
\cluster_1, \cluster_2, \cdots, \cluster_m &= \text{DBSCAN}(\{i \mid \aff_i \ge \text{percentile}_{90}(\aff) \}) \\
\cluster^* &= \arg \max_{\cluster_j} |\cluster_j|
\end{align}

We generate a set of candidate camera positions $\mathcal{P} = \{\pos\}$ via sampling (e.g., farthest point sampling) within the feasible workspace. For each position $\pos$, the camera orientation is constructed by computing the viewing direction $\lookat$ pointing from the camera to the object point cloud centroid $\pcdctr$:
\begin{align}
\lookat &= \frac{\pcdctr - \pos}{\| \pcdctr - \pos \|} \\
\mathbf{r}_x &= \lookat \times \begin{bmatrix}0 & 0 & 1\end{bmatrix}^{\text{T}} \\
\mathbf{R} &= \left[ \frac{\mathbf{r}_x}{\| \mathbf{r}_x \|}, \frac{ \lookat \times \mathbf{r}_x }{\| \lookat \times \mathbf{r}_x \|}, \lookat \right] \in \real{3 \times 3}
\end{align}

Subsequently, we evaluate the predefined candidate set $\mathcal{P}$ in parallel using a weighted loss function $\loss{nbv}$ to directly select the globally optimal view $\pos^*$:
\begin{align}
\loss{nbv}(\pos) &= \loss{orient}(\pos) + w_1 \loss{occ}(\pos) + w_2 \loss{elev}(\pos) \\
\pos^* &= \arg \min_{\pos \in \mathcal{P}} \loss{nbv}(\pos)
\end{align}
Here, the weighting coefficients $w_1$ and $w_2$ balance different sub-objectives and were determined via Bayesian optimization as $w_1 = 0.6$ and $w_2 = 0.2$.

The orientation loss aims to minimize the distance between the camera and the target region. We define $s_i$ as the viewing alignment score for the $i$-th point, calculated as the cosine similarity between the camera view direction $\lookat$ and the vector from point $\pcd_i$ to the centroid:
\begin{align}
s_i &= \lookat^\text{T} \frac{\pcdctr - \pcd_i}{\| \pcdctr - \pcd_i \|} \\
\loss{orient}(\pos) &= 1 - \frac{\sum_{i \in \cluster^*} \aff_i s_i}{\sum_{i \in \cluster^*} \aff_i}
\end{align}

The occlusion loss quantifies occlusion severity by estimating the projected distance of obstacles relative to target points on the image plane. The obstacle point cloud $\pcd^-$ consists of scene points excluding the target region $\cluster^*$. We introduce the angle $\theta_{ij}$ to describe the deviation of an obstacle point $\pcd^-_j$ from the viewing direction, and approximate its projected distance on the image plane as $d_i$. A smaller $d_i$ indicates that the obstacle is aligning closely with the target in the field of view, incurring a heavier occlusion penalty:
\begin{align}
\theta_{ij} &= \arccos \left( \frac{\pcd_i - \pos}{\| \pcd_i - \pos \|} \right)^\text{T} \frac{\pcd_i - \pcd^-_j}{\| \pcd_i - \pcd^-_j \|} \\
d_i &= \min_{j} \| \pcd_i - \pcd^-_j \| \sin \theta_{ij} \\
\loss{occ}(\pos) &= \frac{\sum_{i \in \cluster^*} \aff_i \cdot (1 / (1 + 1000 \cdot d_i))}{\sum_{i \in \cluster^*} \aff_i}
\end{align}
In practice, to ensure real-time performance, we employ a cylindrical query to filter out points far from the line of sight, retaining only nearby obstacle points along the viewing direction for the distance computation above.

The elevation loss prevents the camera from assuming extreme top-down positions. Although such views often offer the largest field of view and highest information gain, relying on them excessively can cause the view planning process to degenerate. This penalty term addresses the issue by suppressing excessive vertical offsets in the camera position:
\begin{align}
\loss{elev}(\pos) &= \frac{|\pos_z|}{\sqrt{\pos_x^2 + \pos_y^2}}
\end{align}

\section{EXPERIMENTS}

\begin{table*}[htb]
  \centering
  \caption{Task-oriented grasping performance with complete shape.}
  \label{tab:full}
  \begin{threeparttable}
  
    \begin{tabular}{lcccccccc}
          \toprule
          \textbf{Method} & \multicolumn{4}{c}{\textbf{Object Instance Generalization}} & \multicolumn{4}{c}{\textbf{Task Generalization}} \\
          \cmidrule(lr){2-5} \cmidrule(lr){6-9}
           & $mAP_{\text{ins}}$ & $mAP_{\text{cls}}$ & $mAP_{\text{task}}$ & $E_{\text{peak}}$ & $mAP_{\text{ins}}$ & $mAP_{\text{cls}}$ & $mAP_{\text{task}}$ & $E_{\text{peak}}$ \\
          \midrule
          
          GCNGrasp-v1~\cite{murali2021SameObjectDifferent} & 79.49 & 76.81 & 73.16 & - & 80.02 & 76.00 & 75.11 & - \\
          GraspGPT*~\cite{tang2023GraspGPTLeveragingSemantic} & 79.70 & \cellcolor{top1}77.88 & 72.84 & - & 79.32 & 76.90 & 72.34 & - \\
          
          \midrule
          
          GCNGrasp-v2 & \cellcolor{top2}80.21 & 77.49 & \cellcolor{top2}74.35 & - & \cellcolor{top1}82.72 & \cellcolor{top1}78.61 & \cellcolor{top1}78.87 & - \\
          + affordance & \cellcolor{top1}80.72 & \cellcolor{top2}77.65 & \cellcolor{top1}74.48 & 0.178 & \cellcolor{top2}82.19 & \cellcolor{top2}77.99 & \cellcolor{top2}76.23 & 0.169 \\
          
          \bottomrule
    \end{tabular}
    
    \begin{tablenotes}
    \footnotesize
        \item[*] indicates data from the original paper.
        \item[\textit{Note}] In this and all subsequent tables, yellow and green backgrounds highlight the \textbf{best} and \textbf{second-best} results, respectively.
    \end{tablenotes}

    \end{threeparttable}
\end{table*}

\begin{table*}[htb]
  \centering
  \caption{Task-oriented grasping performance with partial view.}
  \label{tab:part}
  
    \begin{tabular}{lcccccccc}
      \toprule
      \textbf{Method} & \multicolumn{4}{c}{\textbf{Object Instance Generalization}} & \multicolumn{4}{c}{\textbf{Task Generalization}} \\
      \cmidrule(lr){2-5} \cmidrule(lr){6-9}
       & $mAP_{\text{ins}}$ & $mAP_{\text{cls}}$ & $mAP_{\text{task}}$ & $E_{\text{peak}}$ & $mAP_{\text{ins}}$ & $mAP_{\text{cls}}$ & $mAP_{\text{task}}$ & $E_{\text{peak}}$ \\
      \midrule
      
      GCNGrasp-v1~\cite{murali2021SameObjectDifferent} & 79.28 & 77.08 & 72.78 & - & 79.87 & 76.04 & 74.23 & - \\
      
      \midrule
      
      GCNGrasp-v2 & \cellcolor{top2}80.21 & \cellcolor{top1}77.70 & \cellcolor{top2}74.33 & - & \cellcolor{top2}81.18 & \cellcolor{top2}77.63 & \cellcolor{top1}77.88 & - \\
      + affordance & \cellcolor{top1}80.51 & \cellcolor{top2}77.59 & \cellcolor{top1}75.05 & 0.172 & \cellcolor{top1}81.52 & \cellcolor{top1}77.69 & \cellcolor{top2}76.26 & 0.170 \\
      
      \bottomrule
    \end{tabular}%
\end{table*}

\subsection{Experiment Setup}

All experiments were conducted within a unified computational environment. GCNGrasp-v2 was trained for 200 epochs on two NVIDIA RTX 4090 GPUs, requiring approximately 3 hours. During the testing phase, all TOG models and the view planner performed inference on a single NVIDIA RTX 3090 GPU. This setup ensures experimental consistency while simulating the computational constraints of realistic single-GPU deployments.

We first evaluated TOG performance and affordance prediction quality on the TaskGrasp dataset \cite{murali2021SameObjectDifferent}, covering two settings: object instance generalization and task generalization. Evaluation metrics included mean Average Precision (mAP) for TOG and the relative peak error of the affordance field. Baselines included GCNGrasp \cite{murali2021SameObjectDifferent} and GraspGPT \cite{tang2023GraspGPTLeveragingSemantic}, both trained on the same dataset.

\begin{figure}[htb]
    \centering
    \includegraphics[width=1.0\linewidth]{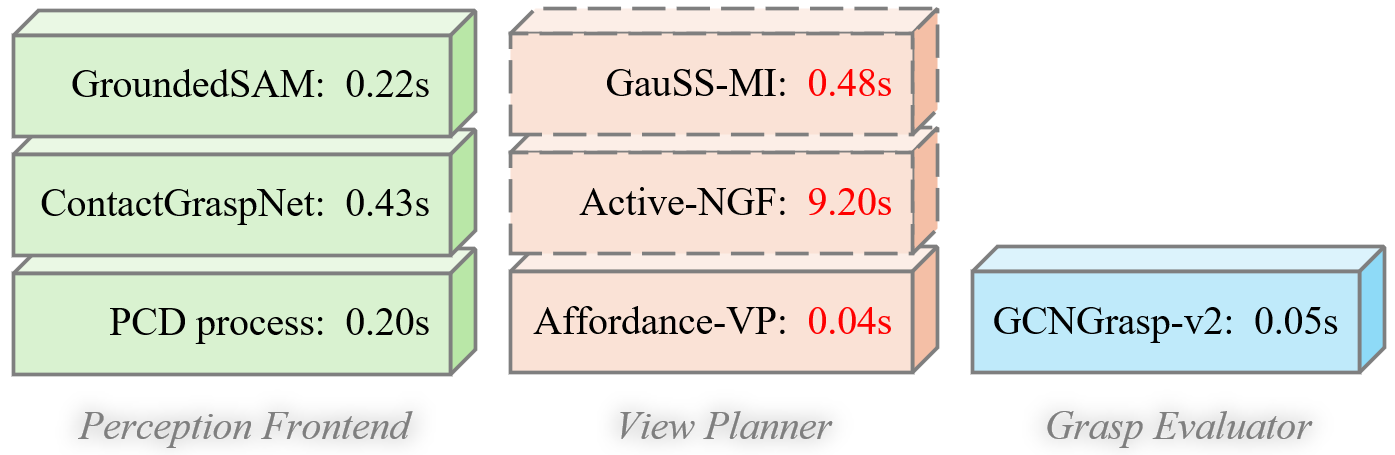}
    \caption{Overview of the experimental system pipeline and per-module inference latency.}
    \label{fig:sys-cost}
\end{figure}

To validate the efficacy of view planning for TOG, we constructed a multi-view observation dataset comprising four object-task pairs. Each scenario includes multi-view RGBD data annotated with TOG ground truths. Furthermore, we employed DepthAnything3~\cite{lin2025DepthAnything3} to perform inter-frame depth alignment, mitigating the impact of sensor noise. As illustrated in Fig.~\ref{fig:sys-cost}, the evaluation system consists of three distinct modules to ensure a fair comparison among different planning methods under identical input features and evaluation criteria:
\begin{itemize}

\item \textbf{Perception Frontend}: Segments target objects via GroundedSAM~\cite{ren2024GroundedSAMAssembling, ravi2024SAM2Segment, liu2025GroundingDINOMarrying} and generates task-agnostic grasp candidates using ContactGraspNet~\cite{sundermeyer2021ContactGraspNetEfficient6DoF}.

\item \textbf{View Planner}: Computes the next best view based on the sequence of historical observations.

\item \textbf{Grasp Evaluator}: Uniformly employs GCNGrasp-v2 to score the task compatibility of candidate grasps generated at each view.
\end{itemize}

Constrained by gravity, the feasible space is restricted to a hemispherical manifold above the object \cite{breyer2022ClosedloopNextbestviewPlanning, shi2025VISOGraspVisionLanguageInformed}. Given its excellent convergence, we opted to validate the approach with few additional views. Under this strategy, experiments commenced from multiple random initial views and seeds, where the camera was sequentially guided to the second and third views. We computed the Average Precision (AP) of predictions against ground truths at each view, with final results reported as the mean AP across all trials. For view planning comparisons, we selected scene-uncertainty-driven methods GauSS-MI~\cite{xie2025GauSSMIGaussianSplatting} and Active-NGF~\cite{jiang2025FisherRFActiveView} as baselines. Leveraging 3D reconstruction capabilities, these methods are theoretically capable of discovering occluded task-relevant regions, representing the state of the art in active perception.

\begin{figure}[htb]
    \centering
    \includegraphics[width=1.0\linewidth]{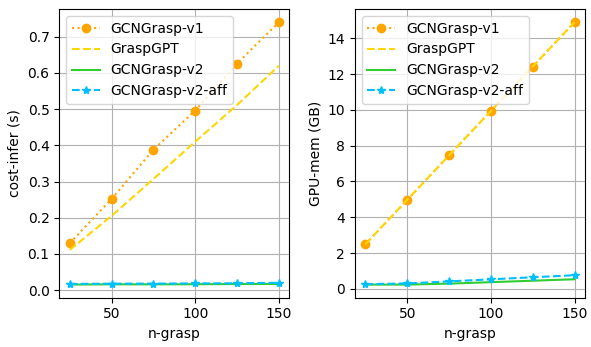}
    \caption{Efficiency comparison of different methods with varying number of grasps.}
    \label{fig:cost-gpu}
\end{figure}

\subsection{Task-Oriented Grasp Evaluation}

TOG prediction accuracy of GCNGrasp-v2 is first evaluated on the TaskGrasp dataset against existing methods \cite{murali2021SameObjectDifferent, tang2023GraspGPTLeveragingSemantic}. Following the protocol in~\cite{murali2021SameObjectDifferent}, inference uses complete object shapes. As shown in Tab.~\ref{tab:full}, the GCNGrasp-v2 series ranks in the top two across the vast majority of metrics and overall outperforms baselines.

Beyond accuracy improvements, GCNGrasp-v2 demonstrates superior computational efficiency. As illustrated in Fig.~\ref{fig:cost-gpu}, inference time and GPU memory consumption of baseline methods grow linearly with the number of candidate grasps. As the number of candidates increases from 25 to 150, baseline inference time rises from approximately 0.1 s to over 0.6 s, while memory usage escalates to nearly 15 GB. In contrast, GCNGrasp-v2 maintains inference time below 0.05 s and memory consumption under 1 GB regardless of candidate count. This constant computational complexity significantly reduces energy consumption and latency, making the model particularly suitable for iterative systems such as view planning that require repeated evaluation of numerous grasp candidates.

In practical scenarios, observations typically begin with single-view partial point clouds. TOG prediction performance under partial views is therefore further evaluated. As shown in Tab.~\ref{tab:part}, performance exhibits only marginal degradation compared to the complete shape setting. This is because the views in the TaskGrasp dataset \cite{murali2021SameObjectDifferent} are relatively ideal, so a single view suffices to provide sufficient cues for the model to make correct decisions.

\subsection{Next Best View Selection}

Tab.~\ref{tab:vp} quantifies the performance of different view planning strategies across four object-task pairs, revealing how view count influences task-oriented grasp prediction accuracy. Initial views often yield suboptimal predictions due to occlusions or unfavorable views, underscoring the necessity of active view selection. While all compared methods improve performance by incorporating additional views, Affordance-VP achieves superior results by precisely focusing on task-relevant regions. Notably, Affordance-VP attains near-saturated prediction performance with only a single view update, significantly reducing perception overhead. A slight performance fluctuation occurs in some tasks when increasing the view count to three, likely attributable to noise accumulation during multi-view feature fusion. Nevertheless, the overall trend demonstrates that our method achieves robust task-oriented grasping predictions with minimal views.

\begin{table}[htb]
    \centering
    \caption{Mean Average Precision (mAP) of task-oriented grasping with varying number of views ($n$).}
    \label{tab:vp}
    
    \begin{tabular}{r l >{\centering\arraybackslash}p{0.8cm} >{\centering\arraybackslash}p{1.6cm} >{\centering\arraybackslash}p{0.8cm} >{\centering\arraybackslash}p{0.8cm}}
        \toprule
        
        $n$ & \textbf{Method} & \multicolumn{1}{c}{\textbf{pan}} & \multicolumn{1}{c}{\textbf{watering can}} & \multicolumn{1}{c}{\textbf{brush}} & \multicolumn{1}{c}{\textbf{cup}} \\
        & & \multicolumn{1}{c}{pour} & \multicolumn{1}{c}{dispense} & \multicolumn{1}{c}{brush} & \multicolumn{1}{c}{drink} \\
        
        \midrule
        
        1 & initial-view & 39.27 & 46.70 & 19.56 & 14.67 \\
        
        \midrule
        
        2 & GauSS-MI~\cite{xie2025GauSSMIGaussianSplatting} & 76.59 & 65.29 & 50.49 & 33.35 \\
        & Active-NGF~\cite{gao2024ActivePerceptionGrasp} & 82.95 & 67.89 & 50.83 & 22.21 \\
        & Affordance-VP & \cellcolor{top1}98.42 & \cellcolor{top2}70.28 & \cellcolor{top2}76.10 & 50.16 \\
        
        \midrule
        
        3 & GauSS-MI~\cite{xie2025GauSSMIGaussianSplatting} & \cellcolor{top2}89.92 & 62.91 & 34.79 & \cellcolor{top2}51.36 \\
        & Active-NGF~\cite{gao2024ActivePerceptionGrasp} & 84.62 & 66.03 & 62.45 & 38.62 \\
        & Affordance-VP & 78.33 & \cellcolor{top1}76.52 & \cellcolor{top1}86.54 & \cellcolor{top1}57.04 \\
                         
        \bottomrule
    \end{tabular}
\end{table}

Fig.~\ref{fig:vis-vp} further visualizes the view planning and grasp prediction results. Taking the brush task as a case study, predicted TOGs in the initial view erroneously concentrate on the bristles rather than the handle. Driven by scene uncertainty, the baselines GauSS-MI~\cite{xie2025GauSSMIGaussianSplatting} and Active-NGF~\cite{jiang2025FisherRFActiveView} prioritize the high geometric entropy of the bristles, neglecting the critical handle region. This misalignment leads to suboptimal view selection and prediction errors. In contrast, Affordance-VP accurately identifies high affordance scores on the handle and actively plans views to directly cover this critical part. This task-semantic-guided strategy avoids the blindness of baselines caused by over-focusing on task-irrelevant regions. 

Real-world experimental results in Tab.~\ref{tab:vp-sr} further validate the effectiveness of the proposed approach. After planning one additional view, Affordance-VP achieves the highest success rates across all four tasks, reaching 100\% in the ``pan pour'' task. In comparison, scene-uncertainty-driven baseline methods exhibit unstable performance in tasks such as ``cup drink'', indicating that their view selection strategies fail to effectively capture critical task-relevant regions. 

\begin{table}[htb]
    \centering
    \caption{Real-world evaluation of view planning for task-oriented grasping. Success rates are reported after executing one planned view movement.}
    \label{tab:vp-sr}
    
    \begin{tabular}{r l >{\centering\arraybackslash}p{0.8cm} >{\centering\arraybackslash}p{1.6cm} >{\centering\arraybackslash}p{0.8cm} >{\centering\arraybackslash}p{0.8cm}}
        \toprule
        
        $n$ & \textbf{Method} & \multicolumn{1}{c}{\textbf{pan}} & \multicolumn{1}{c}{\textbf{watering can}} & \multicolumn{1}{c}{\textbf{brush}} & \multicolumn{1}{c}{\textbf{cup}} \\
        & & \multicolumn{1}{c}{pour} & \multicolumn{1}{c}{dispense} & \multicolumn{1}{c}{brush} & \multicolumn{1}{c}{drink} \\
        
        \midrule
        
        1 & initial-view & 2/24 & 7/20 & 4/28 & 2/24 \\
        
        \midrule
        
        2 & GauSS-MI~\cite{xie2025GauSSMIGaussianSplatting} & 17/24 & 10/20 & 6/28 & 6/24 \\
        & Active-NGF~\cite{gao2024ActivePerceptionGrasp} & 19/24 & 11/20 & 11/28 & 2/24 \\
        & Affordance-VP & \cellcolor{top1}24/24 & \cellcolor{top1}14/20 & \cellcolor{top1}20/28 & \cellcolor{top1}10/24 \\
                         
        \bottomrule
    \end{tabular}
\end{table}

The success rates for TOGs on certain tasks remain suboptimal, primarily due to deviations in predictions of the affordance field. Severe occlusion hinders the network from inferring task-relevance in hidden regions, leading to affordance peaks that deviate from actual grasp locations. This error leads the view planner to select subsequent views with low information gain. Future work will focus on constructing stronger supervision signals to enhance model robustness against incomplete geometric inputs.

Our method also demonstrates significant advantages in computational efficiency. As illustrated in Fig.~\ref{fig:sys-cost}, excluding the necessary preprocessing time of 0.85 s, GCNGrasp-v2 inference and Affordance-VP planning require only 0.05 s and 0.04 s, respectively. In contrast, GauSS-MI~\cite{xie2025GauSSMIGaussianSplatting} and Active-NGF~\cite{gao2024ActivePerceptionGrasp} require 0.48 s and 9.20 s, respectively. This substantial difference in latency stems primarily from the reliance of baseline methods on time-consuming 3D reconstruction processes~\cite{mildenhall2022NeRFRepresentingScenes, kerbl20233DGaussianSplatting}. By operating directly on sparse point clouds and avoiding heavy reconstruction computations, our approach meets the requirements for real-time interaction.

\section{CONCLUSIONS}

This paper presents GCNGrasp-VP, an efficient task-oriented grasping framework that integrates affordance field prediction with view planning to mitigate initial view occlusions. The framework comprises two core components: GCNGrasp-v2, which employs a segmentation-style architecture to enable affordance field prediction with constant-time inference for millisecond-level response; and Affordance-VP, which leverages the affordance field as an information gain metric to drive active view selection toward task-relevant regions without scene reconstruction. 

Experiments demonstrate that our method significantly outperforms scene-uncertainty-driven baselines in view planning tasks, achieving superior performance with only one view adjustment. Real-world validation confirms that the proposed framework substantially improves grasp success rates in single-object scenarios while maintaining minimal computational latency. However, due to inherent deviations in affordance field predictions, our method exhibits limitations in handling certain extreme occlusion scenarios. Future work will focus on constructing stronger supervision signals to bolster the robustness and efficacy of view planning.

\section{Acknowledgment}

This work was supported in part by Shenzhen Science and Technology Program (No. SGDX20240115111759002), in part by Meituan Academy of Robotics Shenzhen, in part by the Shenzhen Association for Science and Technology (No. XHXS2025-003), and in part by High level of special funds (G03034K003) from Southern University of Science and Technology, Shenzhen, China.

\bibliographystyle{IEEEtran}
\bibliography{src/reference}

\addtolength{\textheight}{-12cm}

\end{document}